\documentclass[letterpaper, 10 pt, conference]{ieeeconf}  % Comment this line out if you need a4paper

\IEEEoverridecommandlockouts                              % This command is only needed if 
\usepackage{amsmath} % assumes amsmath package installed
\usepackage{amsfonts}
\usepackage{graphicx}
\usepackage[table]{xcolor} 
\usepackage{multirow}
\usepackage{mathtools}

\usepackage[pagewise]{lineno}
\usepackage{booktabs}
\newcommand{\ra}[1]{\renewcommand{\arraystretch}{#1}}

\usepackage[noadjust]{cite}
\nocite{*}  % Without this, cite articles in text using \cite{...}
\makeatletter
\let\NAT@parse\undefined
\makeatother

\usepackage{hyperref}
\hypersetup{
    colorlinks=true,
    linkcolor=blue,
    filecolor=magenta,      
    urlcolor=blue,
    pdfpagemode=FullScreen,
}
\usepackage{graphicx}
\usepackage{hyperref}
\usepackage{keyval}
\usepackage[ruled,vlined]{algorithm2e}
\usepackage{amsmath}
\usepackage{amssymb}
\usepackage{cleveref}% must be included after amsmath
\usepackage{fancyhdr} 
\usepackage{color}

\newcommand{\idest}{{\it i.e.}, }
\newcommand{\exempli}{{\it e.g.}, }

\definecolor{brick_red}{rgb}{0.80,0.25,0.33}
\definecolor{forest_green}{rgb}{0.2, 0.5, 0.3}
\definecolor{dark_blue}{rgb}{0.1, 0.24, 0.40}
\definecolor{blue_violet}{rgb}{0.2, 0.0, 0.90}
\definecolor{azurean}{rgb}{0.0, 0.3, 0.6}
\definecolor{grey}{rgb}{0.45, 0.45, 0.45}

\newcommand{\scene}{{\color{azurean} \circledtxt{A} \textbf{Scene}}}

\newcommand{\backbone}{{\color{grey} \circledtxt{B} \textbf{Backbone}}}
\newcommand{\locextractor}{{\color{grey} \tinycircledtxt{B1} \textbf{Location Extractor}}}
\newcommand{\vrnn}{{\color{grey} \tinycircledtxt{B2} \textbf{VRNN}}}

\newcommand{\context}{{\color{forest_green} \circledtxt{C} \textbf{Context Module}}}
\newcommand{\patextractor}{{\color{forest_green} \tinycircledtxt{C1} \textbf{Pattern Extractor}}}
\newcommand{\patternnet}{{\color{forest_green} \tinycircledtxt{C2} \textbf{PatternNet}}}
\newcommand{\socinf}{{\color{forest_green} \tinycircledtxt{C3} \textbf{Social Influence}}}
\newcommand{\interactionnet}{{\color{forest_green} \tinycircledtxt{C4} \textbf{InteractionNet}}}

\fancypagestyle{firststyle}
{
  \fancyhf{}
}

\fancypagestyle{secondstyle}
{
  \fancyhf{}
  \fancyhead[R]{\footnotesize \thepage}

}

\newcommand{\update}[1]{\textcolor{black}{#1}}
\newcommand{\sprnn}{\textit{Social-PatteRNN}}
\newcommand{\circledtxt}[1]{\textcircled{\scriptsize{#1}}}
\newcommand{\tinycircledtxt}[1]{\textcircled{\tiny{#1}}}

\title{\LARGE \bf
Social-PatteRNN: Socially-Aware Trajectory Prediction Guided by Motion Patterns
}

\author{Ingrid Navarro$^{1}$ and Jean Oh$^{1}$% <-this % stops a space
\thanks{$^{1}$ Authors are with The Robotics Institute,
     Carnegie Mellon University, 
     {\tt\small \{ingridn, hyaejino\} @andrew.cmu.edu}}%
}

\begin{document}

\maketitle
\thispagestyle{empty}
\pagestyle{empty}

%%%%%%%%%%%%%%%%%%%%%%%%%%%%%%%%%%%%%%%%%%%%%%%%%%%%%%%%%%%%%%%%%%%%%%%%%%%%%%%%
\begin{abstract}
As robots across domains start collaborating with humans in shared environments, algorithms that enable them to reason over human intent are important to achieve safe interplay. 
In our work, we study human intent through the problem of predicting trajectories in dynamic environments. We explore domains where navigation guidelines are relatively strictly defined but not clearly marked in their physical environments. We hypothesize that within these domains, agents tend to exhibit short-term motion patterns that reveal context information related to the agent's general direction, intermediate goals and rules of motion, \exempli social behavior. From this intuition, we propose \textit{Social-PatteRNN}, an algorithm for recurrent, multi-modal trajectory prediction that exploits motion patterns to encode the aforesaid contexts. Our approach guides long-term trajectory prediction by learning to predict short-term motion patterns. It then extracts sub-goal information from the patterns and aggregates it as \textit{social} context.  We assess our approach across three domains: humans crowds, humans in sports and manned aircraft in terminal airspace, achieving state-of-the-art performance.
\end{abstract}

% Original abstract
% As intelligent agents start collaborating with humans in shared space, algorithms that enable them to understand the intent of other agents interacting in a scene are compulsory to achieve seamless and safe interplay. 
%To this end, we propose \textit{Social-PatteRNN}, an algorithm for recurrent, multi-modal trajectory prediction in multi-agent settings. Our approach guides long-term trajectory prediction by learning to predict short-term motion patterns. It then extracts sub-goal information from the patterns and aggregates it as \textit{social} context. 
%We assess its performance across different scenarios: human motion in crowds, human motion in sports and aircraft motion in non-terminal space, and show that our approach generalizes across domains, outperforming state-of-the-art results. 
% and extracting sub-goal information from them while aggregating it as \textit{social} context.  
% Our approach learns short-term motion patterns to guide long-term trajectory prediction, and to extract information about other agents' goals and aggregate it as \textit{social} context. 
%%%%%%%%%%%%%%%%%%%%%%%%%%%%%%%%%%%%%%%%%%%%%%%%%%%%%%%%%%%%%%%%%%%%%%%%%%%%%%%%
\section{Introduction}
Understanding and predicting the intended motion of humans in an environment is an important skill that robots across various domains, \exempli social robotics \cite{ thrun1999minerva, mvrogiannis2021core}, and aerial robotics \cite{patrikar2021trajairnet}, must be equipped with in order to enable safe interactions. Even beyond robotics, other domains such as 
surveillance \cite{oh2011surveillance} or sports analysis \cite{monti2020dagnet, sportvu}
may also benefit from algorithms for modeling intent in dynamic scenes. 

% Nonetheless, 
Motion and intent can be influenced by several factors, making the task challenging. One factor is social behavior, which describes how agents interact, and it %Depending on the context, social behavioral rules may or may not be appropriate.
heavily depends on the context.
For instance, in a sports-based setting, \exempli a basketball game, close proximity to other agents may be a valid behavior, whereas in urban settings, \exempli pedestrians on a side walk, it may not. Respecting motion constraints is another relevant aspect in this setting, as it is related to restrictions that may arise from the agent's own physical constraints, \exempli pedestrians may have more freedom of motion than a vehicle, while vehicles may move at higher speeds, or the constraints imposed by the environment, such as the topology of a scene, or the rules associated with it, \exempli pilots should respect flying guidelines. 
%Few existing works have accounted for agent's dynamic constraints and environmental information \cite{patrikar2021trajairnet, sadeghian2019sophie, salzmann2020trajectron}.
% These may be due to an agent's own motion capabilities or their environmental context 
% \exempli pedestrians have more freedom of motion than a vehicle, while vehicles may move with higher speeds, 
% such as the topology and structure of an environment and the rules associated with it, \exempli vehicles should not drive on the side-walk, or aircraft may need to be aware of their flying altitude.
Moreover, agent motion is driven by each agent's goals which vary depending on the situation and the type of agent: goals can be flexible or fixed, short or long-term, implicit or explicit. This varied nature makes them difficult to model.

\begin{figure}[t]
    \centering
        \centering
         \includegraphics[width=.48\textwidth,trim={0cm 0cm -.8cm -.5cm},clip]{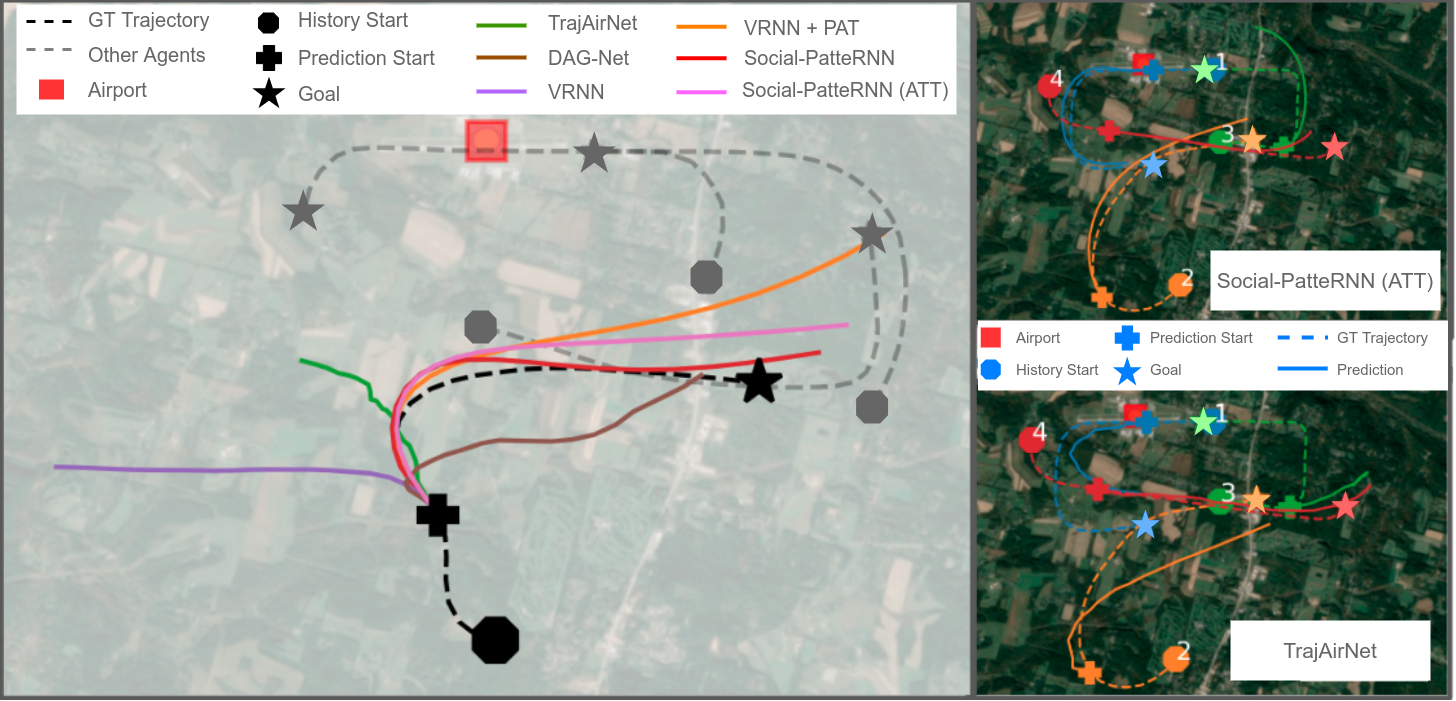}
        
        \caption{Aircraft trajectory predictions. Ground truth (GT) in dashed lines; prediction, solid lines. \textbf{Left:} Various models' predictions vs. GT for an agent. \textbf{Right:} \textit{SocialPatteRNN} (Top) vs. \textit{TrajAirNet} (Bottom) for multiple agents.}%
        \vspace{-0.7cm}%
    \label{fig:trajair_predictions}%
\end{figure}

The majority of existing works on trajectory prediction are mainly focused on pedestrian behaviors~\cite{alahi2016sociallstm,gupta2018socialgan,vemula2018socialattention,zhao2021spec} where the context is relatively loosely defined, \idest except for the social norm that pedestrians try to maintain the comfortable distance from others, there are few rules that guide pedestrian behavior. Moreover, the environmental context  is largely ignored and few existing works, \exempli \cite{sadeghian2019sophie, salzmann2020trajectron} address that. Recent works show promising results by considering the goal context in their predictions~\cite{monti2020dagnet, liu2021avgcn,  manglam2020pecnet}. 

In this paper, we specifically focus on the settings where there exist navigation guidelines relatively strictly defined yet not clearly marked in their physical environments, \exempli in aerial navigation, although there are strict rules analogous to ground navigation, the air space does not display visible lanes. To address this challenge, we propose an approach, known here as \sprnn, where we aim to learn such contexts in the form of motion patterns and social influences extracted from the patterns. Inspired by \cite{zhao2021spec}, we hypothesize that, within these domains, in the short term, agents tend to exhibit some motion patterns that reveal both their general directions of motion and their intermediate goals. We also believe that such patterns further explain general rules of motion, which may be explicit, \exempli sports rules or flying guidelines, or implicit, \exempli social etiquette, as well as, capture admissible motions. From these intuitions, we propose a data-driven approach to learn these patterns of motion and use them as a conditioning signal for predicting multimodal trajectories. Then, we use the learned patterns to extract sub-goal information which we aggregate to our model as the social context.

\begin{figure*}[t!]
    \centering
    \includegraphics[width=1.0\textwidth]{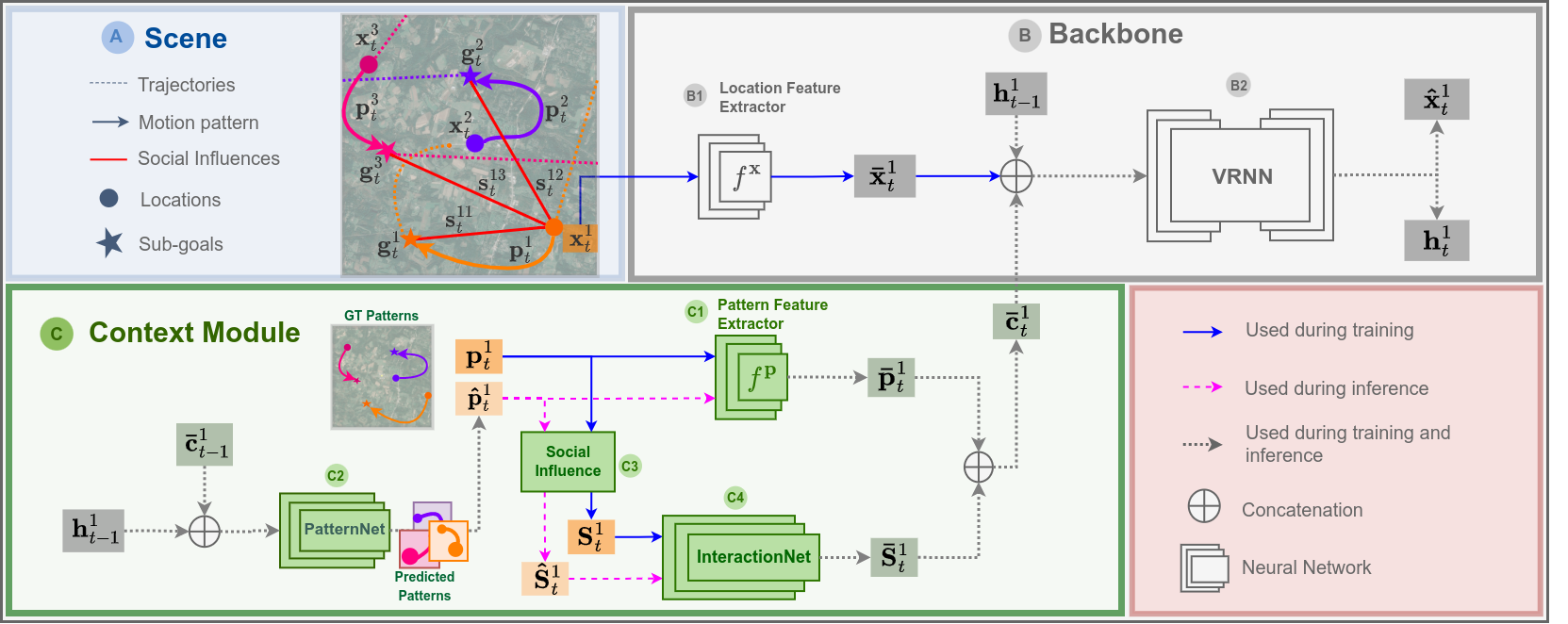}
    \caption{An overview of \sprnn. We show an example of the data flow for {\color{orange}{\textbf{Agent 1}}} during training ({\color{blue}{$\longrightarrow$}}), inference ({\color{magenta}{$\dashrightarrow$}}), and both ({\color{grey}{$\dashrightarrow$}}). Here, \scene~shows an example of a ground truth (GT) scene showing the locations, patterns and sub-goals of all agents at step $t$. 
    \backbone~is the module used for generating locations and summarizing the trajectories. \context~produces the pattern and social features which form the context feature used as input to the backbone.
    %The \patextractor~computes pattern features from GT patterns during training, and from learned patterns during inference. Patterns are learned using \patternnet. Patterns are further used to compute the social influence matrix through the \socinf~module. Finally, \interactionnet~uses the latter matrix to obtain the social features. 
    }
    \label{fig:model}%
    \vspace{-15pt}%
\end{figure*}

To assess the performance and the generalizability of the proposed approach 
%in predicting the trajectories of multiple agents 
across different domains, we evaluate it on three different scenarios: humans in crowds~\cite{robicquet2016sdd,monti2020dagnet}, humans in sports~\cite{monti2020dagnet}, and manned aircraft~\cite{patrikar2021trajairnet}. Experimental results show that the proposed approach has consistent performance across these domains.

Our main contributions are: 1) we propose a generalizable approach for exploiting motion patterns for trajectory prediction; 2) we analyze the proposed approach via ablation to demonstrate the strength of each sub-component; and 3) we share our findings and insights on the sensitivity of the performance to the type of dataset and how it is used, and discuss future directions.
%The paper is organized as follows: Section \ref{sec:related_work} provides details on prior methods for trajectory forecasting. Then, Section \ref{sec:method} formally introduces the approach. Next, Section \ref{sec:experiments} and Section \ref{sec:results} explain implementation and experiment details and show quantitative and qualitative results. Finally, in Section \ref{sec:conclusions}, we briefly discuss conclusions and future directions. 

\section{Related Work}
\label{sec:related_work}

\subsection{Modeling Techniques} 

Recent works tackle socially-aware trajectory prediction through data-driven methods \cite{gupta2018socialgan, alahi2016sociallstm, monti2020dagnet, zhao2021spec}. 
%These algorithms have been vastly explored within the domain of human crowds \cite{gupta2018socialgan, alahi2016sociallstm, monti2020dagnet, zhao2021spec}. 
Here, three main aspects are generally taken into account. First, the use of temporal mechanisms for encoding sequences, %. Two such mechanisms include 
such as Recurrent Neural Networks (RNN) \cite{monti2020dagnet, liu2021avgcn, manglam2020pecnet, alahi2016sociallstm} and Temporal Convolutional Networks (TCN) \cite{nikhill2018tcn, patrikar2021trajairnet, zhao2021spec}, both of which have been shown to exhibit comparable performance \cite{bai2018empirical}. Secondly, modeling the multi-modality of trajectories.
% \idest multiple feasible future trajectories may be valid for a given history. 
For this, researchers have used generative frameworks such as Conditional Variational Autoencoders (C-VAE) \cite{monti2020dagnet, patrikar2021trajairnet, bhattacharyya2019cfvae}, and Generative Adversarial Networks (GAN) \cite{sadeghian2019sophie, zou2018sagail}. 
%Following prior work, we rely on a recurrent C-VAE \cite{chung2015vrnn} to model the aspects above. 
Thirdly, aggregating and modeling social behavior. Common approaches include using social pooling \cite{alahi2016sociallstm, gupta2018socialgan} and attention-based models \cite{monti2020dagnet, liu2021avgcn, yan2021agentformer}. For instance, \cite{alahi2016sociallstm} uses a neighboring-based pooling over the hidden states of a given set of agents. However, as explained in \cite{zhao2021spec}, the drawback to this type of mechanism lies in the difficulty of understanding the semantic representation of hidden states generated by RNNs. Recent works have geared toward using Graph Attention Networks (GAT) to aggregate social context \cite{liu2021avgcn, patrikar2021trajairnet, monti2020dagnet}, showing performance improvements upon prior work. However, as highlighted by \cite{brody2021howattentive, wang2019improvinggat}, GAT models exhibit various limitations such as lack of expressiveness \cite{brody2021howattentive}, and proneness to issues like over-fitting and over-smoothing \cite{wang2019improvinggat}. Other recent approaches have shown promising results using Transformer-based methods \cite{vaswani2017attention} to model and attend over trajectory sequences \cite{yan2021agentformer, giuliari2020tf}. \update{Inspired by these works, we use a self-attention mechanism \cite{vaswani2017attention} to model the dependencies between the social influences of all agents on the current agent (See \Cref{ssec:interaction_net}).}

\subsection{Exploiting Motion Patterns} 

Social Pattern Extraction Convolution (SPEC) \cite{zhao2021spec} also uses motion patterns to learn a pattern scene and predict trajectories from it. Within human crowds, their idea works well because pedestrians tend to move linearly. However, we observed that this model does not generalize to different spatial domains, such as navigating in 3D space, or where motion is highly dynamic, \exempli sports. It also relies on various hyper-parameters for initializing the scene representation and it is sensitive to said initialization. \update{Motivated by this idea, we also leverage motion patterns for trajectory prediction. However, instead of learning a fixed pattern scene, we equip the algorithm with a module that learns to predict patterns from current context information, and then uses them to extract \textit{social} interactions. Our method uses motion patterns to guide the agents' general direction based on the current state, as opposed to enforcing the motion. This is to account for unexpected behavior that may require agents to deviate from the prior intended direction. Finally, our algorithm is not sensitive to initialization and can be easily applied to different spatial domains. 
}
%To address the generalization limitation, rather than assuming an abstract scene representation, we propose to learn to predict patterns of motion from contextual information. 

\section{The Approach}
\label{sec:method}

We consider the problem of predicting the distribution of future trajectories of multiple agents in a shared environment, given the observations of their past trajectories and context information. 
%Here, context refers to the agent's current motion patterns and the social influences extracted from them.
We first formulate this problem as estimating the conditional probability distribution over future trajectories, then, provide the technical details of the proposed approach. 

\subsection{Problem Formulation}
\label{ssec:problem_definition}

The state of an agent $i$ at time step $t$, $\mathbf{x}_{t}^{i}$, represents the coordinates localizing the agent in a scene, \exempli $\mathbf{x}_{t}^{i} = (x, y, z)_{t}^{i}$. Its trajectory is defined as a sequence of $T$ states %from time step $t=1$ to some ending time-step $t=T$, 
\idest $\mathbf{X}^{i} = [\mathbf{x}_{1}^{i}, \mathbf{x}_{2}^{i}, \dots, \mathbf{x}_{T}^{i}]$.
%where $T$ is the length of the trajectory. 
In our approach, we transform the trajectories to relative motions, computed as the displacement between two consecutive steps, $\mathbf{x}_{t}^{i}$ and $\mathbf{x}_{t+1}^{i}$. 
For simplicity, we maintain the same notation for relative coordinates. 

Within this problem setup, a trajectory is split into two segments; a history segment, $\mathbf{X}^{i}_{H} = [\mathbf{x}_{1}^{i}, \dots, \mathbf{x}_{H}^{i}]$ and a future segment, $\mathbf{X}^{i}_{F} = [\mathbf{x}_{H+1}^{i}, \dots, \mathbf{x}_{F}^{i}]$ of lengths $H$ and $F$, respectively. \update{Then, we define the current context, $\mathbf{c}^{i}_t$, and the context history, $\mathbf{C}^i_H = [c^i_1, \dots, c^i_H]$, as information that further describes the agent's current state. For instance, context can be inferred from the observations, learned from previous experiences, or provided from external sources. %(\exempli sensors, maps). 
%Let $\mathbf{c}^{i}_t$ and  $\mathbf{C}^i_H = [c^i_1, \dots, c^i_H]$ denote the context at time step $t$ and the context history, respectively. 
In our approach, context refers to short-term motion patterns as well as social awareness relative to other agents which are both learned from training data.}

The trajectory prediction problem is thus defined as finding the distribution of the future trajectories $\mathbf{\hat{X}}^{i}_{F}$ for each agent $i = \{1, \dots, N\}$ in a scene, given their past trajectories and the context history. Formally,  
\begin{align*} %If not referenced in the rest, use * to remove eq number
    \mathbf{\hat{X}}^{i}_{F} &\sim p(\mathbf{X}^{i}_{F} | \mathbf{X}^{i}_{H}, \mathbf{C}^{i}_{H}) \quad \forall i \in \{1, \dots, N \}.
\end{align*}

\subsection{An Overview of \sprnn}\label{sec:overview}

The proposed approach, known here as \textit{Social-PatteRNN}, is an end-to-end model illustrated in~\Cref{fig:model}. Its \backbone~model consists of a recurrent encoder-decoder model denoted as \vrnn~in the figure, and the \locextractor~for mapping locations into a feature space. 

Our innovations are focused on how context information is learned and used for prediction, \idest in terms of motion patterns and social influences. Specifically, we propose the \context~in~\Cref{fig:model}, consisting of: \patternnet, a module that learns to predict motion patterns given past context which are used to condition trajectory prediction during inference; and \interactionnet~a module that exploits such motion patterns to further reflect \textit{social influences}.
%through the \interactionnet~in~\Cref{fig:model}, which extracts and aggregates \textit{social} context.

Our model can be easily be applied to various spatial navigation domains, \exempli human motions in 2D space or aircraft navigation in 3D. Moreover, its modularity enables to easily incorporate other contextual information.

% Sections \ref{ssec:backbone_model} to \ref{ssec:interaction_net} describe each aspect of the model.

\subsection{Backbone Model}
\label{ssec:backbone_model}

Following prior work \cite{monti2020dagnet, manglam2020pecnet, liu2021avgcn}, we adopt a recurrent version of the C-VAE, \idest the Variational Recurrent Neural Network (VRNN) \cite{chung2015vrnn} for modeling the sequential nature of trajectories and their multi-modality. C-VAEs \cite{sohn2015cvae} are an extension to the VAE \cite{jimenez2014vae}, a type of generative model that is generally well-suited for representing multi-modal distributions by capturing the variations of the input data into a latent representation. C-VAEs \cite{sohn2015cvae} allow to further condition on context variables as a method for controlling the generation process. RNNs capture the sequential dependencies in the data into hidden state variables. 

\update{The VRNN is trained as follows: first it takes a relative location at time-step $t$ for agent $i$, $\mathbf{x}^i_t$ and maps it into a feature space $\mathbf{\bar{x}}^i_t$. It then encodes the location feature along with a context feature $\mathbf{\bar{c}}^i_t$ and the previous hidden state $\mathbf{h}^{i}_{t-1}$ into a latent variable $\mathbf{z}_t^i$ which is then used by a decoder to reconstruct it into $\mathbf{\hat{x}}^i_t$.} This is done for the entire history sequence $\mathbf{X}^i_H$ while also summarizing the hidden representation. During inference, the VRNN receives a history segment as before. Upon processing all history points, it obtains a final hidden state, $\mathbf{h}_H$, which is used to sample a future trajectory sequence of length $F$, $\mathbf{\hat{X}}^i_F$. 

The model consists of four main components, all of which are modeled as neural networks; the encoder (or posterior), $p(\cdot)$, the decoder (or likelihood), $q(\cdot)$, the prior, $p_{\textbf{pr}}(\cdot)$, and the recurrent unit, RNN$(\cdot)$. Additionally, the feature extractor modules for the input data, the latent space and the context vectors, denoted $f^{\mathbf{x}}(\cdot)$, $f^{\mathbf{z}}(\cdot)$, and $f^{\mathbf{c}}(\cdot)$, respectively. In our paper, we denote features with a bar symbol, \exempli $\mathbf{\bar{x}}^i_t$. We note that $f^{\mathbf{z}}(\cdot)$ is implemented within the \vrnn, and we use $f^{\mathbf{c}}(\cdot)$ to simplify notation as it rather refers to the modules introduced in sections \ref{ssec:pattern_net} and \ref{ssec:interaction_net}. 

Mathematically, we can express the model as, 
\begin{align}
    \mathbf{\bar{x}}_t^i &= f^{\mathbf{x}}(\mathbf{x}_{t}^i) \label{eq:fe_x} \\
    \mathbf{\bar{c}}_t^i &= f^{\mathbf{c}}(\mathbf{c}_{t}^i) \label{eq:fe_c} \\
    \mathbf{z}_t^i &\sim p(\mathbf{z}_t^i | \mathbf{\bar{x}}_t^i \oplus \mathbf{\bar{c}}_{t}^i \oplus \mathbf{h}_{t-1}) \quad \text{during training} \label{eq:enc_x}\\
    \mathbf{z}_t^i &\sim p_{\textbf{pr}}(\mathbf{z}_t^i | \mathbf{h}_{t-1}) \quad \text{during inference} \label{eq:prior_x}\\
    \mathbf{\bar{z}}_t^i &= f^{\mathbf{z}}(\mathbf{z}_{t}^i) \label{eq:fe_z} \\
    \mathbf{\hat{x}}_t^i &\sim q(\mathbf{x}_t^i | \mathbf{\bar{z}}_t^i \oplus \mathbf{\bar{c}}_{t}^i \oplus \mathbf{h}_{t-1}) \label{eq:dec_x} \\
    \mathbf{h}_{t}^i &= \text{RNN}(\mathbf{\bar{x}}_t^i \oplus \oplus \mathbf{\bar{z}}_t^i \oplus \mathbf{h}_{t-1}) \label{eq:rnn}
\end{align}
where for all agents $i \in \{1, \dots, N\}$ at a given time step $t$, Eq. \ref{eq:fe_x} and Eq. \ref{eq:fe_c} represent the location and context feature extractor networks, respectively. Eq. \ref{eq:enc_x} and Eq. \ref{eq:prior_x} show the encoder and prior networks, which are needed for learning a latent space $\mathcal{Z}$ and for sampling a corresponding latent representation $\mathbf{z}_t \in \mathcal{Z}$ conditioned on the extracted features and the RNN's hidden states. The encoder is used only during training to guide the prior--which does not have access to the input data--toward the latent space, enabling it to generate future trajectories during the inference stage. The decoder network is expressed in Eq. \ref{eq:dec_x}. It is in charge of reconstructing the location $\mathbf{\hat{x}}_t$ conditioned on the latent variable and context features. The RNN in Eq. \ref{eq:rnn} produces the hidden representation of the current time step using the extracted features and previous state information. 

The loss function for optimizing the VRNN is:
\begin{align*}
    \mathcal{L}^{cvae} &= \sum_{t=1}^{H} \big[ \log{q(\mathbf{x}^i_t | \mathbf{\bar{z}}_t^i \oplus \mathbf{\bar{c}}_{t}^i \oplus \mathbf{h}_{t-1} )} \notag \quad -\\
    &D_{KL}(p(\mathbf{\bar{z}}^i_t | \mathbf{\bar{x}}_t^i \oplus \mathbf{\bar{c}}_{t}^i \oplus \mathbf{h}_{t-1}) \ || \ p_{\mathbf{pr}}(\mathbf{\bar{z}}_t^i |\mathbf{\bar{c}}_{t}^i \oplus \mathbf{h}_{t-1})) \big] \label{eq:loss_elbo}
\end{align*}

The first term represents the log-likelihood for reconstructing the input location while the second is the KL-divergence between the encoder and the prior distributions. For details regarding this equation, we refer the reader to \cite{sohn2015cvae, jimenez2014vae}. 

Distinguished from previous works, we condition the VRNN not only on the input data but also on a context feature which embeds features from short-term motion patterns, $\mathbf{p}^i_t$, and agent-to-agent interactions, $\mathbf{s}^i_t$. 
Sections \ref{ssec:pattern_net} and \ref{ssec:interaction_net} further describe these context features.
% $\mathbf{\bar{c}}^i_t = \mathbf{\bar{p}}^i_t \oplus \mathbf{\bar{s}}^i_t$. The latter embeds features from short-term motion patterns, $\mathbf{p}^i_t$, and agent-to-agent interactions, $\mathbf{s}^i_t$. 
% Sections \ref{ssec:pattern_net} and \ref{ssec:interaction_net} further describe these ideas.
%motion pattern learning and the interaction encoding methods.

\subsection{Learning Motion Patterns}
\label{ssec:pattern_net}

Our approach is based on the intuition that social interactions in navigation are more local (short-term) rather than global (long-term) ones. 
Whereas previous approaches~\cite{monti2020dagnet, patrikar2021trajairnet, manglam2020pecnet} use the agents' long-term goals as the context to guide the learning process, we use short-term motion patterns to learn rules of motion, \exempli \textit{social} behavior (Section \ref{ssec:interaction_net}). 
%Additionally, we hypothesize that these patterns reveal further context information, such as sub-goals and admissible motions. 

In our work, a motion pattern consists of a short trajectory sequence of length $P$, where $P$ is a hyperparameter.  A pattern can be represented as absolute coordinates, relative displacements, or controls such as accelerations. Here, we represent it as a sequence of relative displacements from a given location $\mathbf{x}^i_t$, \idest $\mathbf{p}^{i}_t = [\mathbf{x}^i_t, \dots, \mathbf{x}^i_{t+P}]$. \scene~in \Cref{fig:model} depicts an example of the pattern representation.

We obtain pattern features, $\mathbf{\bar{p}}^i_t$, from motion patterns using the \patextractor~in \Cref{fig:model}. Mathematically, 
\begin{align}
    \mathbf{\bar{p}}^i_t &= f^{\mathbf{p}}(\mathbf{p}^{i}_t)
\end{align}
As explained in prior sections, the pattern features form part of the context feature vector used for training the VRNN.  
%using  pattern information by extracting features from pattern sequences in ground truth data. We refer to these features as $\mathbf{\bar{p}}^i_t$ and to their corresponding feature extractor as $f^{\mathbf{p}}$ which is part of the context feature extraction module  defined in Eq. \ref{eq:fe_c} and shown in \Cref{fig:model} - \textcircled{\tiny{C1}}.

During inference, since we aim to generate new data, we do not have direct access to pattern information. To address this, we train a neural network to predict future motion patterns given the previous pattern information and hidden state. We refer to this network as \patternnet,
\begin{align}
    \mathbf{\hat{p}}^i_t &= PatternNet(\mathbf{\bar{p}}^i_{t-1}, \mathbf{h}^i_{t-1}) \label{eq:patternet}
\end{align}

We optimize this network through a mean squared error (MSE) reconstruction loss,
\begin{align*}
    \mathcal{L}^{pat} = \sum_{t=1}^H \text{MSE}(\mathbf{p}^i_t, \mathbf{\hat{p}}^i_t)
\end{align*}

\subsection{Learning Social Interactions}
\label{ssec:interaction_net}

In addition to using pattern features to guide the trajectory prediction process, we are interested in aggregating context relating to the interactions between agents, which we refer to as social influences. For this, we take advantage of our predicted patterns to extract short-term sub-goal information from all agents in a scene. To do so, for an agent $i$ in the scene, we deem the endpoint of its predicted pattern as their next sub-goal, $\mathbf{g}^i_t$. Then, we compute the \socinf~as the displacement between its current location $\mathbf{x}^i_t$ and its next sub-goal as well as the sub-goals of all other agents. We denote these influences as $\mathbf{s}^{ij}_t$, where $i$ and $j$ are the indices of the agents used to calculate the displacement, \idest $\mathbf{s}^{ij}_t = \mathbf{g}^{j}_t - \mathbf{x}^{i}_t$. The full social influence is $\mathbf{S}^i_t = [\mathbf{s}^{i1}_t, \dots, \mathbf{s}^{iN}_t]$ which we use to extract social context features,
\begin{align}
    \mathbf{\bar{S}}^i_t &= f^{\mathbf{S}}(\mathbf{S}^{i}_t)
    \label{eq:f_d}
\end{align}
% As mentioned before, these features are also part of the context feature vector.  

\update{Since the final social influence vector is computed relative to each agent, we additionally incorporate a multi-head self-attention mechanism to further extract the relationships between the individual influences of all agents on the current agent and get a final social feature which captures these relationships. We refer the reader to \cite{vaswani2017attention} for further details of this mechanism. Here, we refer to the latter feature extractor with self-attention as \interactionnet,
}
\begin{align}
    \mathbf{\bar{S}}^i_t &= InteractionNet(\mathbf{S}^i_{t-1}) \label{eq:interactionnet}
\end{align}
% To further capture the relationships between agents' sub-goals, we use multi-head attention (MHA) \cite{vaswani2017attention} to attend over the displacement features and get a final social context vector $\mathbf{\bar{s}}^i_t$. In this process, sub-goal information is projected into $h$ input queries $\mathbf{q}$, keys $\mathbf{k}$, and values $\mathbf{v}$ (Eq. \ref{eq:qkv}) and then transformed into $h$ independent attention heads (Eq. \ref{eq:heads}). These heads are then combined together and linearly transformed to produce the attended social context vector (Eq. \ref{eq:cat}), 
% \begin{align}
%     [(\mathbf{q}, \mathbf{k}, \mathbf{v})_1, \dots, (\mathbf{q}, \mathbf{k}, \mathbf{v})_h] &= f^{\mathbf{qkv}}(\mathbf{d}^{i}_t) \label{eq:qkv} \\
%     [\text{head}_1, \dots, \text{head}_h] &= \text{MHA}([\mathbf{q}, \mathbf{k}, \mathbf{v}]_{i=1}^h) \label{eq:heads} \\
%     \mathbf{\bar{s}}^i_t &= f^{\mathbf{d}}(\text{cat}([\text{head}_1, \dots, \text{head}_h])) \label{eq:cat}
% \end{align}
% \begin{align}
%     \mathbf{\bar{s}}^i_t &= InteractionNet(\mathbf{d}^i_{t-1}) \label{eq:interactionnet}
% \end{align}

% In \Cref{sec:experiments}, we show the effect of using context features with (Eq. \ref{eq:f_d}) and without (Eq. \ref{eq:cat}) the attention mechanism.

Now, we can finally define the context feature vector introduced in~\Cref{ssec:problem_definition} as $\mathbf{\bar{c}}^i_t = \mathbf{\bar{p}}^i_t \oplus \mathbf{\bar{S}}^i_t$. As shown in~\Cref{fig:model}, context features are fed to the VRNN backbone. We also feed the context vector to the previously introduced \patternnet. Thus, Eq. \ref{eq:patternet} now becomes,
\begin{align*}
    \mathbf{\hat{p}}^i_t &= PatternNet(\mathbf{\bar{c}}^i_{t-1}, \mathbf{h}^i_{t-1})
\end{align*}

Finally, the full loss function for optimizing our model is,
\begin{align*}
    \mathcal{L}^{total} = \mathcal{L}^{pat} + \mathcal{L}^{cvae}
\end{align*}

\label{ssec:proposed_approach}

\section{Experiments}
\label{sec:experiments}

\subsection{Datasets}

We evaluate our model on the following datasets:
% 1) aircraft in non-towered airspace, 2) crowds, and 3) sports.
% Below, we describe the datasets used in each domain.

\subsubsection{\textbf{TrajAir}} TrajAir \cite{patrikar2021trajairnet} is a dataset designed for studying social navigation in non-towered airspace. It consists of 111 days of manned aircraft trajectory data. In \cite{patrikar2021trajairnet}, the authors train their model on a subset consisting of 28 days of data. We consider the 111 days of data; but, we partition the dataset differently. Since the dataset is heavily unbalanced toward single agent scenes (\Cref{tab:trajair_statistics}), and we are interested in modeling social behavior in scenes with multiple aircraft, we train our models on a subset where scenes contain at least 4 agents. 

\begin{table}[h]
\centering
\caption{Number of train/test samples (in thousands) by number of agents per scene using trajectories of $T=140$s.}
\ra{1.3}
\resizebox{\columnwidth}{!}{
\begin{tabular}{@{}ccccccc@{}}
\toprule
\# agents & all & 1 & 2 & 3 & 4 & +4\\
\midrule 
\# trajectories & 425 / 177 & 298 / 177 & 90 / 42 & 26 / 14 & 9 / 3 & 2 / 0.2 \\
% All & $\sim$425k / $\sim$ 177k \\
\bottomrule
\end{tabular}%
}%
% \vspace{-8pt}
\label{tab:trajair_statistics}%
\end{table}

\subsubsection{\textbf{Stanford Drone Dataset (SDD)}} SDD \cite{robicquet2016sdd} is dataset consisting of top-down videos capturing social behavior in complex and crowded scenarios with different types of agents, \exempli pedestrians, bikers, skateboarders, and vehicles.
%as well as environmental complexity including physical structures such as roundabouts. 
We use the version in~\cite{monti2020dagnet} where only pedestrians are considered. 

\subsubsection{\textbf{SportVU NBA Dataset (NBA)}} NBA \cite{sportvu} is a dataset collected with a SportVU system~\cite{sportvu} consisting of motion tracking data of 10 basketball players; 5 attackers and 5 defenders. We use the version of NBA as in \cite{monti2020dagnet}. In \cite{monti2020dagnet}, models are trained on each team separately as they consider the nature of the teams to be intrinsically different. We train on both teams jointly as we argue that agent behaviors influence the interactions within their own team as well as the opponent team.

%\subsection{Evaluation Methodology}

\subsection{Evaluation} \label{ssec:evaluation}
Following prior work on trajectory prediction \cite{alahi2016sociallstm, gupta2018socialgan, zhao2021spec, manglam2020pecnet, patrikar2021trajairnet, monti2020dagnet, liu2021avgcn, vemula2018socialattention}, we consider Average Displacement Error (ADE) and Final Displacement Error (FDE) as our metrics for assessing performance of our models. We report the best ADE and FDE values of $K=20$ samples.

\update{We first evaluate our approach by ablating its components. Specifically, we validated the backbone consisting of the VRNN, the use of motion patterns without social context (\textit{PAT)}, aggregating social interactions (\textit{SOC}), and finally, attending over the interactions (\textit{ATT}). Then, we} evaluate \sprnn~against four related models: \textit{A-VRNN} \cite{monti2020dagnet}, \textit{DAG-Net} \cite{monti2020dagnet}, \textit{SPEC} \cite{zhao2021spec}, and \textit{TrajAirNet} \cite{patrikar2021trajairnet}. We adapted \textit{DAG-Net} and \textit{SPEC} for 3D space, and \textit{TrajAirNet} for 2D. 

\subsection{Implementation Details}

We used the \textit{VRNN} in \cite{monti2020dagnet}. The feature extractors as well as \textit{PatternNet} were implemented as fully-connected networks. For the attention mechanism in the \textit{InteractionNet} we used 4 attention heads. We use Adam optimizer and train for at most 1000 epochs with and learning rate of $1e-3$ and early stopping. For the \textit{SDD} and \textit{NBA} datasets, we followed \cite{monti2020dagnet}, where trajectories are split into $H=8$ and $F=12$, and $H=10$ and $F=40$, respectively. Empirically, we set the pattern legths for \textit{SDD} to $P=6$ and for \textit{NBA} to $P=8$. For \textit{TrajAir}, we use trajectory segments of 140 seconds and split them into $H=40$ and $F=100$ sampled every 5 steps, and pattern lengths of $P=20$s \cite{patrikar2021trajairnet}. 

% \subsubsection{\textbf{Training}}
% Our experiments were trained using Adam optimizer for at most 600 epochs and performing early stopping. We used mini-batches of size 64 and learning rate of $1e-4$. 

\section{Results}

\subsection{Ablation Study} 
\update{As explained in \cref{ssec:evaluation}, we validated the benefit of each component of the model by performing ablations. The corresponding results, summarized in \Cref{tab:ablation_experiments}, show that using the pattern learning module improves the results in term of displacement errors in the three datasets. Moreover, exploiting pattern information to aggregate \textit{social} features, achieves further reductions in error. We note that the more pronounced improvements correspond to \textit{TrajAir}'s results, where, adding pattern learning reduces ADE / FDE by $\sim12/9\%$, while aggregating social features further improves these metrics by  $\sim5\%$. We find this specially relevant since in aircraft navigation, pilots are expected to follow flying patterns in addition to procuring safe interactions with other pilots \cite{patrikar2021trajairnet}. We also observe that in the sports domain, where interactions are more frequent, aggregating social encoding achieves more significant improvements in ADE / FDE, of  $\sim11/14\%$, than only considering motion patterns, $\sim3/2\%$. 
}

\subsection{Main Results}
\Cref{fig:trajair_predictions} shows a qualitative result for the \textit{TrajAir} dataset. 
%For ease of visualization we only show the predictions for the agent with the blue trajectory. (in the caption)
In this example, we can observe that the predicted trajectories for \textit{TrajAirNet} and the \textit{VRNN} baseline do not align with the correct direction of the ground truth trajectory, while our models that were conditioned on additional context do. We can also observe that while the end-point of the trajectory predicted by \textit{DAG-Net} gets close to the goal location, it does not follow the expected aircraft pattern, whereas our pattern-conditioned approaches do. \update{Furthermore, our pattern-based models achieve smoother predictions than \textit{TrajAirNet} and \textit{DAG-Net}. We believe that these observations support our initial hypothesis that through motion patterns additional context information can be learned, \exempli general direction and rules of motion.
Quantitatively, \Cref{tab:result_comparison} summarizes the results for the \textit{socially}-aware baselines introduced in \Cref{ssec:evaluation} and ours, which we show as \sprnn and \sprnn~\textit{(ATT)}, for the models without and with attention, respectively. From these results we observe that our models outperform existing baselines in both \textit{TrajAir} and \textit{NBA}, and achieves comparable performance in the \textit{SDD} dataset w.r.t to the state-of-the-art model, \textit{DAG-Net}.}
As an additional finding, we observe that the performances of \textit{SPEC} and \textit{DAG-Net}, developed for human data, degrades on the aerial navigation dataset, while \textit{TrajAirNet}, developed for aerial navigation, falls behind on human datasets.

In \Cref{fig:trajair_predictions}, we can also see the benefit of aggregating social context information by comparing the predicted trajectory by our approaches \textit{with} social context (\sprnn~and \textit{Social-PatteRNN (ATT)}), shown in pink and red, respectively,  and that \textit{without} social context (VRNN-PAT), shown in orange.
Quantitatively, \Cref{tab:ablation_experiments} concurs that adding social context further improves performance. 

\label{sec:results}
\begin{table}[h]
\centering
\caption{The ablation results in ADE / FDE ($\downarrow$).}
\ra{1.3}
\resizebox{\columnwidth}{!}{\begin{tabular}{@{}ll cccccc@{}}
\toprule
& & & \textbf{Trajair (km)} & & \textbf{SDD (m)} & & \textbf{NBA (ft)} \\
\midrule 
1 & VRNN & & 0.660 / 1.392 & & 0.605 / 1.181 & & 9.176 / 14.375  \\
\midrule
2 & \quad + PAT & & 0.580 / 1.264 & & 0.587 / 1.177 & & 8.877 / 14.110 \\
% 3 & \quad + \todo{PatternNet (L)} &   &  &  &  & & & \\
% % \midrule
3 & \quad + PAT + SOC & & 0.555 / 1.203 & & \textbf{0.552} / \textbf{1.099} & & 8.312 / 12.604  \\
% % 5 & \quad + \todo{PatternSNet (L)} &   &  &  &  &  & & \\
% \midrule
4 & \quad + PAT + SOC + ATT & & \textbf{0.551} / \textbf{1.192} & & 0.565 / 1.118 & & \textbf{8.125} / \textbf{12.342}  \\
% % 7 & \quad + \todo{PatternSNet (L) + MHA} &   &  &  &  &  & & \\
\bottomrule
\end{tabular}}
\label{tab:ablation_experiments}
\end{table}

\begin{table}[h]
\centering
\caption{Model comparison in ADE / FDE ($\downarrow$).}
\ra{1.3}
\resizebox{\columnwidth}{!}{\begin{tabular}{@{}ll cccccc@{}}
\toprule
& & & \textbf{Trajair (km)} & & \textbf{SDD (m)} & & \textbf{NBA (ft)} \\
\midrule 
1 & A-VRNN \cite{monti2020dagnet} & & 0.64 / 1.31 & & 0.56 / 1.14 & & 8.88 / 14.06  \\
2 & DAG-Net \cite{monti2020dagnet} & & 0.78 / 1.53 & & \textbf{0.53} / \textbf{1.04} & & 8.55 / 12.37  \\
3 & TrajAirNet \cite{patrikar2021trajairnet} & & 0.77 / 1.50 & & 0.92 / 1.68 & & 9.91 / 15.27  \\
4 & S-PEC \cite{zhao2021spec} & & 0.96 / 2.05 & & 0.70 / 1.11 & & 11.03 / 12.54 \\
\midrule
5 & Social-PatteRNN (Ours) & & 0.56 / 1.20 & & 0.55 / 1.10 & & 8.31 / 12.60  \\
6 & Social-PatteRNN-ATT (Ours) & & \textbf{0.55} / \textbf{1.19}& & 0.57 / 1.12 & & \textbf{8.13} / \textbf{12.34}  \\
\bottomrule
\end{tabular}}
\label{tab:result_comparison}
\end{table}

\section{CONCLUSIONS}
\label{sec:conclusions}

In our work, we hypothesized that short-term motion patterns reveal context information related to the agents' general direction, rules of motion, and how they interact with each other. From this intuition, we introduced \sprnn, a trajectory prediction algorithm that leverages motion patterns to capture these contexts. 
% general direction of motion and to extract and aggregate social context relating to the interactions between agents. 
%One of our main motivations for this work is to explore methods that are able to generalize across domains. As such, we showed that our approach generalizes across three different scenarios, human crowds, sports and aircraft navigation outperforming previous state-of-the-art. 
Our experiments show that the performance of existing approaches are sensitive to problem domain and how the dataset is composed. The results support our intuition that short-term patterns are robust indicators for trajectory prediction and are also less sensitive to problem domains. As future work, we aim to explore various avenues. We are interested in further exploiting and learning the dependencies between the agents by exploring other attention mechanism, as well as other backbone structures such as Transformers \cite{yan2021agentformer}. We are also interested in incorporating more context information from  input modalities other than trajectory data. Finally, we plan to extend our work to develop a safe and socially-aware robot navigation algorithm for urban driving and aerial navigation in crowded scenarios. 
%are interested in applying our model to other domains, \eg vehicles in urban settings, and exploring additional techniques for incorporating useful context for trajectory prediction. 

% This command serves to balance the column lengths
% on the last page of the document manually. It shortens
% the textheight of the last page by a suitable amount.
% This command does not take effect until the next page
% so it should come on the page before the last. Make
% sure that you do not shorten the textheight too much.
\addtolength{\textheight}{0cm}  

\section*{ACKNOWLEDGMENT}

% \scriptsize{
We thank Jay Patrikar for their valuable feedback. This work was supported, in part, by the Army Futures Command Artificial Intelligence Integration Center (AI2C) and the Ministry of Trade, Industry and Energy (MOTIE) and Korea Institute of Advancement of Technology (KIAT) through the International Cooperative R\&D program: P0019782, Embeded AI Based fully autonomous driving software and Maas technology development. Views expressed in here do not necessarily represent those of the aforementioned entities.
% }

\bibliography{refs}

\end{document}